\documentclass[11pt,a4paper]{article}
\usepackage[hyperref]{naaclhlt2019}
\usepackage{times}
\usepackage{latexsym}
\usepackage{multicol}
\usepackage{multirow}
\usepackage{algorithm}
\usepackage[noend]{algpseudocode}
\usepackage[colorinlistoftodos]{todonotes}
\usepackage{comment}
\usepackage{enumitem}
\usepackage{breqn}
\usepackage{tabularx}
\usepackage{arydshln}
\usepackage{graphicx}

\usepackage{array}
\newcolumntype{L}[1]{>{\raggedright\let\newline\\\arraybackslash\hspace{0pt}}m{#1}}
\newcolumntype{C}[1]{>{\centering\let\newline\\\arraybackslash\hspace{0pt}}m{#1}}
\newcolumntype{R}[1]{>{\raggedleft\let\newline\\\arraybackslash\hspace{0pt}}m{#1}}

\usepackage{url}

\aclfinalcopy % Uncomment this line for the final submission

\title{Complexity-Weighted Loss and Diverse Reranking \\ for Sentence Simplification}

\author{\parbox{12cm}{\centering Reno Kriz$^{*}$, Jo\~ao Sedoc$^{*}$, Marianna Apidianaki$^{\triangle}$, \\ Carolina Zheng$^{*}$, Gaurav Kumar$^{*}$, Eleni Miltsakaki$^{\dagger}$, \\ and Chris Callison-Burch$^{*}$} \\
$^{*}$ Computer and Information Science Department, University of Pennsylvania \\
$^{\triangle}$ LIMSI, CNRS, Universit\'e Paris-Saclay, 91403 Orsay \& LLF, Univ. Paris Diderot \\
$^{\dagger}$ Choosito, Inc. \\ 
{\tt \{rekriz,joao,gauku,carzheng,ccb\}@seas.upenn.edu}, \\ {\tt marianna@limsi.fr, eleni@choosito.com}
}

\date{}

\begin{document}
\maketitle

\begin{abstract}
    Sentence simplification is the task of rewriting texts so they are easier to understand. Recent research has applied sequence-to-sequence (Seq2Seq) models to this task, focusing largely on training-time improvements via reinforcement learning and memory augmentation. One of the main problems with applying generic Seq2Seq models for simplification is that these models tend to copy directly from the original sentence, resulting in outputs that are relatively long and complex. We aim to alleviate this issue through the use of two main techniques. 
    First, we incorporate content word complexities, as predicted with a leveled word complexity model, into our loss function during training.
    Second, we generate a large set of diverse candidate simplifications at test time, and rerank these to promote fluency, adequacy, and simplicity. Here, we measure simplicity through a novel sentence complexity model. 
    These extensions allow our models to perform competitively with state-of-the-art systems while generating simpler sentences.
    We report standard automatic and human evaluation metrics.\footnote{Our code is available in our fork of Sockeye \cite{hieber2017sockeye} at https://github.com/rekriz11/sockeye-recipes.} 
\end{abstract}

\section{Introduction}

Automatic text simplification aims to reduce the complexity of texts and preserve their meaning, making their content more accessible to a broader audience \cite{saggion2017automatic}.
This process can benefit people with reading disabilities, foreign language learners and young children, and can assist non-experts exploring a new field. Text simplification has gained wide interest in recent years due to its relevance for NLP tasks. Simplifying text during preprocessing can improve the performance of syntactic parsers \cite{chandrasekar1996motivations} and semantic role labelers \cite{vickrey2008sentence,woodsend2014text}, and can improve the grammaticality (fluency) and meaning preservation (adequacy) of  translation output \cite{stajner2016text}.

\begin{figure}[bt]
    \centering
\includegraphics[width=7.5cm]{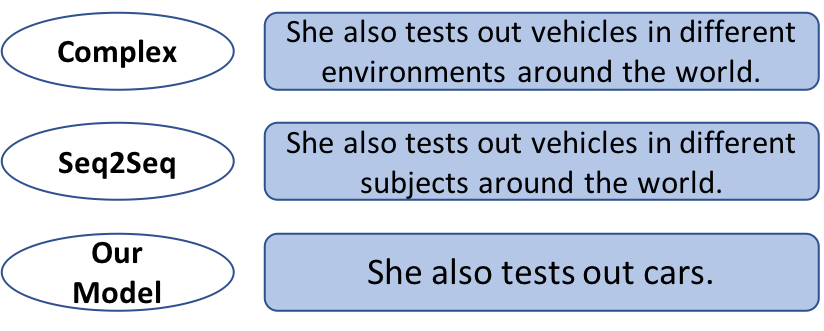}
\caption{Example comparison of a simplification generated by a standard Seq2Seq model vs. our model.}
\label{example}
\end{figure}

Most text simplification work has approached the task as a monolingual machine translation problem \cite{woodsend2011learning,narayan2014hybrid}. Once viewed as such, a natural approach is to use sequence-to-sequence (Seq2Seq) models, which have shown state-of-the-art performance on a variety of NLP tasks, including machine translation \cite{vaswani2017attention} and dialogue systems \cite{vinyals2015neural}.

One of the main limitations in applying standard Seq2Seq models to simplification is that these models tend to copy directly from the original complex sentence too often, as this is the most common operation in simplification. Several recent efforts have attempted to alleviate this problem using reinforcement learning \cite{zhang2017sentence} and memory augmentation \cite{zhao2018integrating}, but these systems often still produce outputs that are longer than the reference sentences. To avoid this problem, we propose to extend the generic Seq2Seq framework at both training and inference time by encouraging the model to choose simpler content words, and by
effectively choosing an output based on a large set of 
candidate simplifications. The main contributions of this paper can be summarized as follows:
\begin{itemize}
    \item We propose a custom loss function to replace standard cross entropy probabilities during training, which takes into account the complexity of content words.
    \item We include a similarity penalty at inference time to generate more diverse simplifications, and we further cluster similar sentences together to remove highly similar candidates.
    \item We develop methods to rerank candidate simplifications to promote fluency, adequacy, and simplicity, helping the model choose the best option from a diverse set of sentences.
\end{itemize}

An analysis of each individual components reveals that of the three contributions, reranking simplifications at post-decoding stage brings about the largest benefit for the simplification system. We compare our model to several state-of-the-art systems in both an automatic and human evaluation settings, and show that the generated simple sentences are shorter and simpler, while remaining competitive with respect to fluency and adequacy. We also include a detailed error analysis to explain where the model currently falls short and provide suggestions for addressing these issues.

\section{Related Work}

Text simplification has often been addressed as a monolingual translation process, which generates a simplified version of a complex text. \newcite{zhu2010monolingual} employ a tree-based translation model and consider sentence splitting, deletion, reordering, and substitution. \newcite{coster2011learning} use a Phrase-Based Machine Translation (PBMT) system with support for deleting phrases, while  \newcite{wubben2012sentence} extend a PBMT system with a reranking heuristic (PBMT-R). \newcite{woodsend2011learning} propose a model based on a quasi-synchronous grammar, a formalism able to capture structural mismatches and complex rewrite operations. \newcite{narayan2014hybrid} combine a sentence splitting and deletion model with PBMT-R. This model has been shown to perform competitively with neural models on automatic metrics, though it is outperformed using human judgments \cite{zhang2017sentence}.

In recent work, Seq2Seq models are widely used for sequence transduction tasks such as machine translation \cite{sutskever2014sequence,luong2015effective}, conversation agents \cite{vinyals2015neural}, summarization \cite{nallapati2016abstractive}, etc. Initial Seq2Seq models consisted of a Recurrent Neural Network (RNN) that encodes the source sentence $\textbf{x}$ 
to a hidden vector of a fixed dimension, followed by another RNN that uses this hidden representation to generate the target sentence $\textbf{y}$. The two RNNs are then trained jointly to maximize the conditional probability of the target sentence given the source sentence, i.e. $P(\textbf{y}|\textbf{x})$. Other works have since extended this framework to include attention mechanisms \cite{luong2015effective} and transformer networks \cite{vaswani2017attention}.\footnote{For a detailed description of Seq2Seq models, please see \cite{sutskever2014sequence}.} 
%Nisioi et al.'s 
\newcite{nisioi2017exploring} 
%work 
was the first major application of Seq2Seq models to text simplification, applying a standard encoder-decoder approach with attention and beam search.
\newcite{vu2018sentence} extended this framework to incorporate memory augmentation, which simultaneously performs lexical and syntactic simplification, allowing them to outperform standard Seq2Seq models.

There are two main Seq2Seq models we will compare to in this work, along with the statistical model from \newcite{narayan2014hybrid}. \newcite{zhang2017sentence} proposed DRESS (Deep REinforcement Sentence Simplification), a Seq2Seq model that uses a reinforcement learning framework at training time to reward the model for producing sentences that score high on fluency, adequacy, and simplicity. This work showed state-of-the-art results on human evaluation. However, the sentences generated by this model are in general longer than the reference simplifications. \newcite{zhao2018integrating} proposed DMASS (Deep Memory Augmented Sentence Simplification), a multi-layer, multi-head attention transformer architecture which also integrates simplification rules. 
This work has been shown to get state-of-the-art results in an automatic evaluation, training on the WikiLarge dataset introduced by \newcite{zhang2017sentence}.
\newcite{zhao2018integrating}, however, does not perform a human evaluation, and restricting evaluation to automatic metrics is generally insufficient for comparing simplification models. 
Our model, in comparison, is able to generate shorter and simpler sentences according to Flesch-Kincaid grade level \cite{kincaid1975derivation} and human judgments, and provide a comprehensive analysis using human evaluation and a qualitative error analysis.

\section{Seq2Seq Approach} \label{methods}

\subsection{Complexity-Weighted Loss Function} \label{loss}

Standard Seq2Seq models use cross entropy as the loss function at training time. This only takes into account how similar our generated tokens are to those in the reference simple sentence, and not the complexity of said tokens. Therefore, we first develop a model to predict word complexities, and incorporate these into a custom loss function.

\subsubsection{Word Complexity Prediction} \label{word-comp-section}

Extending the complex word identification model of \newcite{kriz2018simplification}, we train a linear regression model using length, number of syllables, and word frequency; we also include Word2Vec embeddings \cite{mikolov2013distributed}. To collect data for this task, we consider the Newsela corpus, a collection of 1,840 news articles written by professional editors at 5 reading levels \cite{xu2015problems}.\footnote{Newsela is an education company that provides reading materials for students in elementary through high school. The Newsela corpus can be requested at https://newsela.com/data/} We extract word counts in each of the five levels; in this dataset, we denote 4 as the original complex document, 3 as the least simplified re-write, and 0 as the most simplified re-write. We propose using Algorithm \ref{alg:word-complexity} to obtain the complexity label for each word $w$, where $l_w$ represents the level given to the word, and $c_{w_i}$ represents the number of times that word occurs in level $i$.

\begin{algorithm}
\caption{Word Complexity Data Collection}
\label{alg:word-complexity}

\begin{algorithmic}[1]
\Procedure{Data Collection}{}
\State $l_w \gets 4$ 
\For{$i \in \{3, 0\}$}
    \If{$c_{w_i} \geq 0.7*c_{w_{i+1}}$}
        \If{$c_{w_i} \geq 0.4*c_{w_4}$}
            \State $l_w \gets i$
        \EndIf
    \EndIf
\EndFor

\Return $l_w$
\EndProcedure
\end{algorithmic}
\end{algorithm}

Here, we initially label the word with the most complex level, 4. If at least 70\% of the instances of this word is preserved in level 3, we reassign the label as level 3; if the label was changed, we then do this again for progressively simpler levels.

\begin{table}
\begin{center}
\begin{tabular}{|c|c|c|} \hline
\textbf{Model} & \textbf{Correlation} & \textbf{MSE} \\ \hline
Frequency Baseline & -0.031 & 1.90 \\
Length Baseline & 0.344 & 1.51 \\
LinReg & \textbf{0.659} & \textbf{0.92} \\ \hline
\end{tabular}
\end{center}
\caption{\label{word-comp} Pearson Correlation and Overall Mean Squared Error (MSE) of the word-level complexity prediction model (LinReg). Comparison to length-based and frequency-based baselines.}
\end{table}

As examples, Algorithm \ref{alg:word-complexity} labels ``pray", ``sign", and ``ends" with complexity level 0, and ``proliferation", ``consensus", and ``emboldened" with complexity level 4. We split the data extracted from Algorithm \ref{alg:word-complexity} into Train, Validation and Test sets (90\%, 5\% and 5\%, respectively, and use them for training and evaluating the complexity prediction model. \footnote{Note that we also tried continuous rather than discrete labels for words by averaging frequencies, but found that this increased the noise in the data. For example, ``the" and ``dog" were incorrectly labeled as level 2 instead of 0, since these words are seen frequently across all levels.}

We report the Mean Squared Error (MSE) and Pearson correlation on our test set in Table \ref{word-comp}.\footnote{We report MSE results by level in the appendix.} We compare our model to two baselines, which predict complexity using log Google $n$-grams frequency \cite{thorsten2006web} and word length, respectively. For these baselines, we calculate the minimum and maximum values for words in the training set, and then normalize the values for words in the test set.

\subsubsection{Loss Function}

We propose a metric that modifies cross entropy loss to upweight simple words while downweighting more complex words. More formally, the probabilities of our simplified loss function can be generated by the process described in Algorithm \ref{alg:loss-function}. Since our word complexities are originally from 0 to 4, with 4 being the most complex, we need to reverse this ordering and add one, so that more complex words and non-content words are not given zero probability. In this algorithm, we denote the original probability vector as \textbf{CE}, our vocabulary as $\textbf{V}$, the predicted word complexity of a word $v$ as $score_v$, the resulting weight for a word as $w_v$, and our resulting weights as \textbf{SCE}, which we then normalize and convert back to logits.

\begin{algorithm}
\caption{Simplified Loss Function}
\label{alg:loss-function}

\begin{algorithmic}[1]
\Procedure{Simplified Loss}{}
\State $\textbf{CE} \gets$ softmax($logits_{CE}$)
\For{$v \in \textbf{V}$}
    \State $score_v \gets WordComplexity(v)$
    \If{$v$ is a content word}
        \State $w_v \gets (4 - s_v) + 1$
    \Else
        \State $w_v \gets 1$
    \EndIf
\EndFor
\State $w_v \gets \Big(\frac{w_v}{\sum_{v\in V}w_v}\Big)^{\alpha}$ for $v \in \textbf{V}$
\State $\textbf{SCE} \gets \textbf{CE} \cdot \textbf{w}$

\Return $\textbf{SCE}$
\EndProcedure
\end{algorithmic}
\end{algorithm}

Here, $\alpha$ is a parameter we can tune during experimentation.
Note that we only upweight simple content words, not stopwords or entities.

\subsection{Diverse Candidate Simplifications} \label{diverse}

To increase the diversity of our candidate simplifications, we apply a beam search scoring modification proposed in \newcite{li2016simple}. In standard beam search with a beam width of $b$, given the $b$ hypotheses at time $t-1$, the next set of hypotheses is generated by first selecting the top $b$ candidate expansions from each hypothesis. These $b\times b$ hypotheses are then ranked by the joint probabilities of their sequence of output tokens, and the top $b$ according to this ranking are chosen.

We observe that candidate expansions from a single parent hypothesis tend to dominate the search space over time, even with a large beam. To increase diversity, we apply a penalty term based on the rank of a generated token among the $b$ candidate tokens from its parent hypothesis.

If $Y^j_{t-1}$ is the $j^{th}$ top hypothesis at time $t-1$, $j\in[1..b]$, and $y_t^{j,j'}$ is a candidate token generated from $Y^j_{t-1}$, where $j'\in[1..b]$ represents the rank of this particular token among its siblings, then our modified scoring function is as follows (here, $\delta$ is a parameter we can tune during experimentation):
\begin{equation}
\small
S(Y_{t-1}^j,y_t^{j,j'})=\log{p(y_1^j,\dots,y_{t-1}^j,y_t^{j,j'}|x)}-j'*\delta
\end{equation}

Extending the work of \newcite{li2016simple}, to further increase the distance between candidate simplifications, we can cluster similar sentences after decoding. To do this, we convert each candidate into a document embedding using Paragraph Vector \cite{le2014distributed}, cluster the vector representations using $k$-means, and select the sentence nearest to the centroids. This allows us to group similar sentences together, and only consider candidates that are relatively more different.

\subsection{Reranking Diverse Candidates}

Generating diverse sentences is helpful only if we are able to effectively rerank them in a way that promotes simpler sentences while preserving fluency and adequacy. To do this, we propose three ranking metrics for each sentence $i$:

\begin{table}
\begin{center}
\begin{tabular}{|c|c|c|} \hline
\textbf{Model} & \textbf{Correlation} & \textbf{MSE} \\ \hline
Length Baseline & 0.503 & 3.72 \\
CNN (ours) & \textbf{0.650} & \textbf{1.13} \\ \hline
\end{tabular}
\end{center}
\caption{\label{sent-comp-table} Pearson Correlation and Overall Mean Squared Error (MSE) for the sentence-level complexity prediction model (CNN), compared to a length-based baseline.}
\end{table}

\begin{itemize}
    \item \textbf{Fluency} ($f_i$): We calculate the perplexity based on a 5-gram language model trained on English Gigaword v.5 \cite{parker2011english} using KenLM \cite{heafield2011kenlm}.
    \item \textbf{Adequacy} ($a_i$): We generate Paragraph Vector representations \cite{le2014distributed} for the input sentence and each candidate and calculate the cosine similarity.
    \item \textbf{Simplicity} ($s_i$): We develop a sentence complexity prediction model to predict the overall complexity of each sentence we generate.
\end{itemize}

To calculate sentence complexity, we modify a Convolutional Neural Network (CNN) for sentence classification \cite{kim2014convolutional} to make continuous predictions. We use aligned sentences from the Newsela corpus \cite{xu2015problems} as training data, labeling each with the complexity level from which it came.\footnote{We respect the train/test splits described in Section \ref{simpdata}.} As with the word complexity prediction model, we report MSE and Pearson correlation on a held-out test set in Table \ref{sent-comp-table}.\footnote{We report MSE results by level in the appendix.}

We normalize each individual score between 0 and 1, and calculate a final score as follows:

\begin{equation}
    score_i = \beta_ff_i + \beta_aa_i + \beta_ss_i
\end{equation}

\noindent We tune these weights (\textbf{$\beta$}) on our validation data during experimentation to find the most appropriate combinations of reranking metrics. Examples of improvements resulting from the including each of our contributions are shown in Table \ref{examples}.

\section{Experiments}

\subsection{Data} \label{simpdata}

We train our models on the Newsela Corpus. In previous work, models were mainly trained on the parallel Wikipedia corpus (PWKP) consisting of paired sentences from English Wikipedia and Simple Wikipedia \cite{zhu2010monolingual}, or the extended WikiLarge corpus \cite{zhang2017sentence}. We choose to instead use Newsela, because it was found that 50\% of the sentences in Simple Wikipedia are either not simpler or not aligned correctly, while Newsela has higher-quality simplifications \cite{xu2015problems}.

As in \newcite{zhang2017sentence}, we exclude sentence pairs corresponding to levels 4-3, 3-2, 2-1, and 1-0, where the simple and complex sentences are just one level apart, as these are too close in complexity. After this filtering, we are left with 94,208 training, 1,129 validation, and 1,077 test sentence pairs; these splits are the same as \newcite{zhang2017sentence}. We preprocess our data by tokenizing and replacing named entities using CoreNLP \cite{manning2014stanford}. 

\begin{table*}
\begin{center}
\begin{tabular}{| L{4cm} | c | L{3.5cm} | c | L{3cm} |} \hline
\textbf{Complex Sentence} & \textbf{Model 1} & \textbf{Model 1 Sentence} & \textbf{Model 2} & \textbf{Model 2 Sentence} \\ \hline

{\small Mary travels between two offices.} & S2S & {\small Mary is a professor at the park.} & S2S-Loss & {\small Mary goes between two offices.} \\ \hline

{\small Their fatigue changes their voices, but they're still on the freedom highway.} & S2S & {\small Their condition changes their voices, but they're still on the freedom highway.} & S2S-FA & {\small Their fatigue changes their voices.} \\ \hline

{\scriptsize Just until recently, the education system had banned Islamic headscarves in schools and made schoolchildren recite a pledge of allegiance.} & S2S-FA & {\small The education system had banned Islamic law.} & S2S-Cluster-FA & {\small Only until recently , the education system had banned Islamic hijab in schools.} \\ \hline

{\small Police used tear gas, dogs and clubs on the unarmed protesters.} & S2S-FA & {\small Police used tear gas and dogs on the unarmed protesters.}& S2S-Diverse-FA & {\small They used tear gas and dogs.}\\ \hline
\end{tabular}
   \caption{\label{examples}Example sentences where each component of our model improved the output sentence, compared to a model that does not use that component.}
\end{center}
\end{table*}

\subsection{Training Details}

For our experiments, we use Sockeye, an open source Seq2Seq framework built on Apache MXNet \cite{hieber2017sockeye,chen2015mxnet}. In this model, we use LSTMs with attention for both our encoder and decoder models with 256 hidden units, and two hidden layers. We attempt to match the hyperparameters described in \newcite{zhang2017sentence} as closely as possible; as such, we use 300-dimensional pretrained GloVe word embeddings \cite{pennington2014glove}, and Adam optimizer \cite{kingma2015adam} with a learning rate of 0.001. We ran our models for 30 epochs.\footnote{All non-default hyperparameters can be found in the Appendix.}

During training, we use our complexity-weighted loss function, with $\alpha = 2$; for our baseline models, we use cross-entropy loss. At inference time, where appropriate, we set the beam size $b = 100$, and the similarity penalty $\delta = 1.0$. After inference, we set the number of clusters to 20, and we compare two separate reranking weightings: one which uses fluency, adequacy, and simplicity (FAS), where $\beta_f = \beta_a = \beta_s = \frac{1}{3}$; and one which uses only fluency and adequacy (FA), where $\beta_f = \beta_a = \frac{1}{2}$ and $\beta_s$ = 0.

\subsection{Baselines and Models}

We compare our models to the following baselines:

\begin{itemize}
    \item \textbf{Hybrid} performs sentence splitting and deletion before simplifying with a phrase-based machine translation system \cite{narayan2014hybrid}.
    \item \textbf{DRESS} is a Seq2Seq model trained with reinforcement learning which integrates lexical simplifications \cite{zhang2017sentence}.\footnote{For Hybrid and DRESS, we use the generated outputs provided in \newcite{zhang2017sentence}. We made a significant effort to rerun the code for DRESS, but were unable to do so.}
    \item \textbf{DMASS} is a Seq2Seq model which integrates the transformer architecture and additional simplifying paraphrase rules \cite{zhao2018integrating}.\footnote{For DMASS, we ran the authors' code on our data splits from Newsela, in collaboration with the first author to ensure an accurate comparison.}
\end{itemize}

We also present results on several variations of our models, to isolate the effect of each individual improvement.
\textbf{S2S} is a standard sequence-to-sequence model with attention and greedy search. \textbf{S2S-Loss} is trained using our complexity-weighted loss function and greedy search. \textbf{S2S-FA} uses beam search, where we rerank all sentences using fluency and adequacy (FA weights). \textbf{S2S-Cluster-FA}  clusters the sentences before reranking using FA weights. \textbf{S2S-Diverse-FA} uses diversified beam search, reranking using FA weights. \textbf{S2S-All-FAS} uses all contributions, reranking using fluency, adequacy, and simplicity (FAS weights). 
Finally, \textbf{S2S-All-FA} integrates all modifications we propose, and reranks using FA weights.

\section{Results} \label{results}

In this section, we compare the baseline models and various configurations of our model with both standard automatic simplification metrics and a human evaluation. We show qualitative examples where each of our contributions improves the generated simplification in Table \ref{examples}.

\begin{table}
\begin{center}
\begin{tabular}{|l|c|c|} \hline
\textbf{Model} & \textbf{SARI} & \textbf{Oracle} \\ \hline
Hybrid & 33.27 & -- \\
DRESS & 36.00 & -- \\
DMASS & 34.35 & -- \\ \hline
S2S & 36.32 & -- \\
S2S-Loss & 36.03 & -- \\
S2S-FA & 36.47 & 54.01 \\
S2S-Cluster-FA & \textbf{37.22} &  50.36 \\
S2S-Diverse-FA & 35.36 & 52.65 \\
S2S-All-FAS & 36.30 & 50.40 \\
S2S-All-FA & \textbf{37.11} & 50.40 \\ \hline
\end{tabular}
\end{center}
\caption{\label{sari} Comparison of our models to baselines and state-of-the-art models using SARI. We also include oracle SARI scores (Oracle), given a perfect reranker. S2S-All-FA is significantly better than the DMASS and Hybrid baselines using a student t-test ($p < 0.05$).}
\end{table}

\subsection{Automatic Evaluation}

Following previous work \cite{zhang2017sentence,zhao2018integrating}, we use SARI as our main automatic metric for evaluation \cite{xu2016optimizing}.\footnote{To calculate SARI, we use the original script provided by \cite{xu2016optimizing}.} Specifically, SARI calculates how often a generated sentence correctly keeps, inserts, and deletes $n$-grams from the complex sentence, using the reference simple standard as the gold-standard, where $1 \leq n \leq 4$. Note that we do not use BLEU \cite{papineni2002bleu} for evaluation; even though it correlates better with fluency than SARI, \newcite{sulem2018bleu} recently showed that BLEU often negatively correlates with simplicity on the task of sentence splitting. We also calculate oracle SARI, where appropriate, to show the score we could achieve if we had a perfect reranking model. Our results are reported in Table \ref{sari}.

Our best models outperform previous state-of-the-art systems, as measured by SARI. Table \ref{sari} also shows that, when used separately, reranking and clustering result in improvements on this metric. Our loss and diverse beam search methods have more ambiguous effects, especially when combined with the former two; note however that including diversity before clustering does slightly improve the oracle SARI score.

We calculate several descriptive statistics on the generated sentences and report the results in Table \ref{stats}. We observe that our models produce sentences that are much shorter and lower reading level, according to Flesch-Kincaid grade level (FKGL) \cite{kincaid1975derivation}, while making more changes to the original sentence, according to Translation Error Rate (TER) \cite{snover2006study}. In addition, we see that the customized loss function increases the number of insertions made, while both the diversified beam search and clustering techniques individually increase the distance between sentence candidates.

\begin{table}
\begin{center}
\small
\begin{tabular}{|l|ccccc|} \hline
\textbf{Model} & \textbf{Len} & \textbf{FKGL} & \textbf{TER} & \textbf{Ins} & \textbf{Edit}\\ \hline
Complex & 23.1 & 11.14 & 0 & 0 & -- \\ \hline
Hybrid & 12.4 & 7.82 & 0.49 & 0.01 & --\\
DRESS & 14.4 & 7.60 & 0.44 & 0.07 & --\\ 
DMASS & 15.1 & 7.40 & 0.59 & 0.28 & -- \\ \hline
S2S & 16.1 & 7.91 & 0.41 & 0.23 & -- \\
S2S-Loss & 16.4 & 8.11 & 0.40 & 0.31 & --\\
S2S-FA & 7.6 & 6.42 & 0.73 & 0.01 & 7.28 \\
{\scriptsize S2S-Cluster-FA} & 9.1 & 6.49 & 0.68 & 0.05 & 7.55 \\
{\scriptsize S2S-Diverse-FA} & 7.5 & 5.97 & 0.78 & 0.07 & 8.22 \\
S2S-All-FAS & 9.1 & 5.37 & 0.68 & 0.05 & 7.56 \\
S2S-All-FA & 10.8 & 6.42 & 0.61 & 0.07 & 7.56 \\ \hline
Reference & 12.8 & 6.90 & 0.67 & 0.42 & --\\
\hline
\end{tabular}
\end{center}
\caption{\label{stats} Average sentence length, FKGL, TER score compared to input, and number of insertions. We also calculate average edit distance (Edit) between candidate sentences for applicable models.}
\end{table}

\begin{comment}
\begin{table}
\begin{center}
\small
\begin{tabular}{|l|cccccc|} \hline
{\normal{\textbf{Model}}} & {\normal{\textbf{SLen}}} & {\normal{\textbf{WLen}}} & {\normal{\textbf{FKGL}}} & {\normal{\textbf{TER}}} & {\normal{\textbf{Ins}}} & {\normal{\textbf{Edit}}}\\ \hline
Hybrid & 12.4 & 4.41 & 7.82 & 0.49 & 0.01 & --\\
DRESS & 14.4 & 4.23 & 7.60 & 0.44 & 0.07 & --\\ 
DMASS & 15.1 & 4.08 & 7.40 & 0.59 & 0.28 & -- \\ \hline
S2S & 16.1 & 4.17 & 7.91 & 0.41 & 0.23 & -- \\
S2S-Loss & 16.4 & 4.20 & 8.11 & 0.40 & 0.31 & --\\
S2S-FA & 7.6 & 4.28 & 6.42 & 0.73 & 0.01 & 7.28 \\
{\scriptsize S2S-Cluster-FA} & 9.1 & 4.22 & 6.49 & 0.68 & 0.05 & 7.55 \\
{\scriptsize S2S-Diverse-FA} & 7.5 & 4.17 & 5.97 & 0.78 & 0.07 & 8.22 \\
S2S-All-FAS & 9.1 & 4.06 & 5.37 & 0.68 & 0.05 & 7.56 \\
S2S-All-FA & 10.8 & 4.22 & 6.42 & 0.61 & 0.07 & 7.56 \\ \hline
Reference & 12.8 & 4.24 & 6.9 & 0.67 & 0.42 & --\\
\hline
\end{tabular}
\end{center}
\caption{\label{stats} Average sentence (SLEN) and word length (WLEN), FKGL, TER score compared to input, and number of insertions. We also calculate average Edit Distance (Edit) between sentences before reranking, for applicable models.}
\end{table}
\end{comment}

\begin{table*}
\begin{center}
\begin{tabular}{|l|cccc|}
\hline
\textbf{Model} & \textbf{Fluency} & \textbf{Adequacy} & \textbf{Simplicity} & \textbf{All} \\ \hline
Hybrid & 2.79* & 2.76 & 2.88* & 2.81* \\
DRESS & \textbf{3.50} & \textbf{3.11}* & 3.03 & \textbf{3.21}* \\ 
DMASS & 2.59* & 2.15* & 2.50* & 2.41* \\ \hline
S2S-All-FAS & 3.35 & 2.50* & \textbf{3.11} & 2.99 \\
S2S-All-FA & 3.38 & 2.66 & \textbf{3.08} & 3.04 \\ \hline
Reference & 3.82* & 3.23* & 3.29* & 3.45* \\ \hline
\end{tabular}
\end{center}
\caption{\label{likert1} Average ratings of crowdsourced human judgments on fluency, adequacy and complexity. Ratings significantly different from S2S-All-FA are marked with * ($p < 0.05$); statistical significance tests were calculated using a student t-test. We provide 95\% confidence intervals for each rating in the appendix.}
\end{table*}

\subsection{Human Evaluation}

While SARI has been shown to correlate with human judgments on simplicity, it only weakly correlates with judgments on fluency and adequacy \cite{xu2016optimizing}. Furthermore, SARI only considers simplifications at the word level, while we believe that a simplification metric should also take into account sentence structure complexity. We plan to investigate this further in future work.

Due to the current perceived limitations of automatic metrics, we also choose to elicit human judgments on 200 randomly selected sentences to determine the relative overall quality of our simplifications. For our first evaluation, we ask native English speakers on Amazon Mechanical Turk to evaluate the fluency, adequacy, and simplicity of sentences generated by our systems and the baselines, similar to \newcite{zhang2017sentence}. Each annotator rated these aspects on a 5-point Likert Scale. These results are found in Table \ref{likert1}.\footnote{We present the instructions for all of our human evaluations in the appendix.}

As we can see, our best models substantially outperform the Hybrid and DMASS systems. Note that DMASS performs the worst, potentially because the transformer model is a more complex model that requires more training data to work properly. Comparing to DRESS, our models generate simpler sentences, but DRESS better preserves the meaning of the original sentence.

To further investigate why this is the case, we know from Table \ref{stats} that sentences generated by our model are overall shorter than other models, which also corresponds to higher TER scores. \newcite{napoles2011evaluating} notes that on sentence compression, longer sentences are perceived by human annotators to preserve more meaning than shorter sentences, controlling for quality. Thus, the drop in human-judged adequacy may be related to our sentences' relatively short lengths.

To test that this observation also holds true for simplicity, we took the candidates generated by our best model, and after reranking them as before, we selected three sets of sentences:

\begin{itemize}
    \item \textbf{MATCH-Dress0}: Highest ranked sentence with length closest to that of DRESS (DRESS-Len); average length is 14.10.
    \item \textbf{MATCH-Dress+2}: Highest ranked sentence with length closest to (DRESS-Len + 2); \\ average length is 15.32.
    \item \textbf{MATCH-Dress-2}: Highest ranked sentence with length closest to (DRESS-Len - 2); \\ average length is 12.61.\\
\end{itemize}

\begin{figure}[bt]
    \centering
\includegraphics[width=7.5cm]{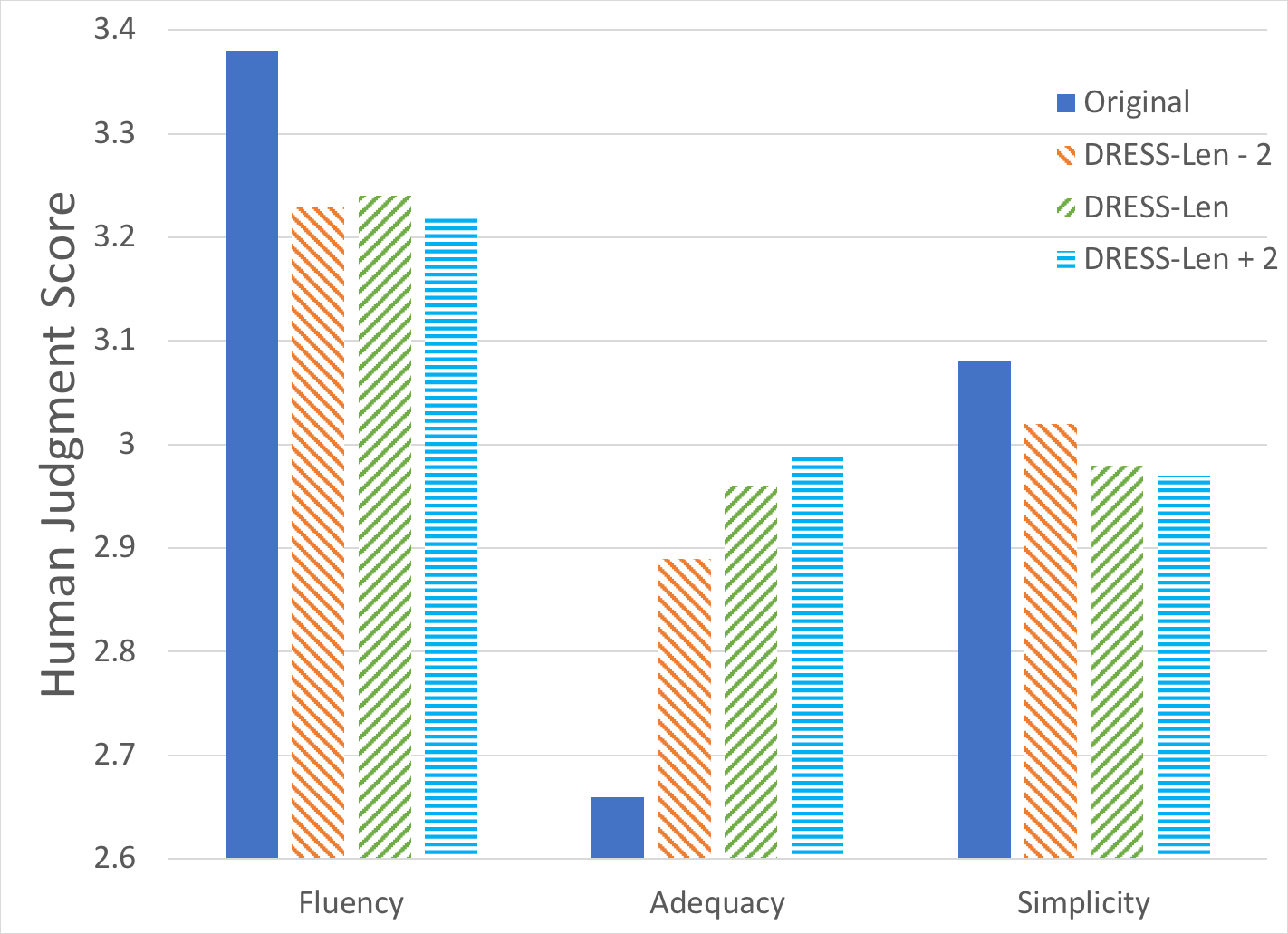}
\caption{Effect of length on human judgments.}
\label{length}
\end{figure}

The average fluency, adequacy, and simplicity from human judgments on these new sentences are shown in Figure \ref{length}, along with those ranked highest by our best model (Original). As expected, meaning preservation does substantially increase as we increase the average sentence length, while simplicity decreases. Interestingly, fluency also decreases as sentence length increases; this is likely due to our higher-ranked sentences having greater fluency, as defined by language model perplexity.

\section{Error Analysis} 
To gain insight in what aspects of the simplification process are challenging to our model, we present the most recurring types of errors from our test set. \\

\subsection{Examples of Error Types}
\begin{enumerate}
\item{Long and complex sentences with multiple clauses}

\begin{enumerate}
\small
\item{\label{longandcomplex1}} \underline{\it Complex}: Turkey has long enshrined the secular ideals of founding father Mustafa Kemal Ataturk, particularly in an education system that until recently banned Islamic headscarves in schools and made schoolchildren begin the day reciting an oath of allegiance to Ataturk's legacy. \\
\underline{\it Reference}: Schools in Turkey had banned headscarves.\\
\underline{\it Simple}: They made schoolchildren to Ataturk's history.

\item{\label{longandcomplex2}} \underline{\it Complex}: And Wal-Mart, which imports more fruits and vegetables from Mexico than any other U.S. company, announced its effort to force improvements up and down its supply chain.\\
\underline{\it Reference}: Experts said Wal-Mart is an important company.\\
\underline{\it Simple}: Wal-Mart used more fruits and vegetables from the company.
\end{enumerate}

\item{Need for anaphora resolution}
\begin{enumerate}
\small
\item {\label{anaphoraresolution1}} \underline{\it Complex}: He is the creative director of Rethink Leisure \& Entertainment , which is working on several projects in China and elsewhere in Asia . \\
\underline{\it Reference}: He is with Rethink Leisure \& Entertainment.\\
\underline{\it Simple}: He is working on several projects in China.

\item {\label{anaphoraresolution2}} \underline{\it Complex}: Teachers there say Richie reads like a high school student.\\
\underline{\it Reference}: He reads like a high school student. \\
\underline{\it Simple}: Richie says he is a high school student.
\end{enumerate}

\item{Simplifying the wrong part of the sentence}
\begin{enumerate}
\small
\item{\label{wrongpart1}} \underline{\it Complex}: Parks deliberately maintained her image as shy and proper, said Adrienne Cannon, an expert on African-American history.\\
\underline{\it Reference}: Adrienne Cannon studies African-American history.\\
\underline{\it Simple}: She is an expert on African-American history.

\item{\label{wrongpart2}}\underline{\it Complex}: His father owned the home when the lava flowed slowly to the coast.\\
\underline{\it Reference}: His father still owned the home.\\
\underline{\it Simple}: The river cut slowly to the coast.\\
\end{enumerate}

\item{Poor substitution due to word embeddings proximity}
\begin{enumerate}
\small
\item{\label{badsub1}} \underline{\it Complex}: Just until recently, the education system had banned Islamic \textbf{headscarves} in schools and made schoolchildren begin the day reciting a pledge of allegiance to Ataturk's legacy.\\
\underline{\it Reference}: Schools in Turkey had banned headscarves.\\
\underline{\it Simple}: Just until recently, the education officials had banned  Islamic \textbf{flags}.
\item {\label{badsub2}} \underline{\it Complex}: In Beijing kite circles, Fei is widely known as the elder \textbf{statesman}.\\
\underline{\it Reference}: In Beijing, Fei is widely known as an expert.\\
\underline{\it Simple}: In Beijing, Fei is considered a \textbf{doctor}.
\end{enumerate}

\item{Training confusion due to misalignments in training corpus}
\begin{enumerate}
\small
\item {\label{badalignment1}} \underline{\it Complex}: He heard the applause.\\
\underline{\it Reference}: The winning word was ``magician."\\
\underline{\it Simple}: But he heard the song.

\item{\label{badalignment2}} \underline{\it Complex}: More than 300,000 students across the country signed up to participate. \\
\underline{\it Reference}: Rogers Middle School took part in the program.\\
\underline{\it Simple}: There were more than 300,000 students.
\end{enumerate}

\item{Acceptable simplifications that were not available in the training}
\begin{enumerate}
\small
\item{\label{acceptable1}} \underline{\it Complex}: He later tried to buy his freedom.\\
\underline{\it Reference}: Henson tried to buy his freedom later. \\
\underline{\it Simple}: The man tried to buy his freedom.

\item{\label{acceptable2}} \underline{\it Complex}: Middle school was a rough couple of years for Talmus' daughter, Lili Smith. \\
\underline{\it Reference}: She had a hard time in middle school.\\
\underline{\it Simple}: School was a rough couple of years.
\end{enumerate}
\end{enumerate}

\subsection{Error Discussion}

Attempting to rewrite very long and complex sentences resulted to consistent errors, as shown in \ref{longandcomplex1} and \ref{longandcomplex2}. This observation in combination with the examples of mis-alignments in the training corpus (\ref{badalignment1} and \ref{badalignment2}) indicate that we either need to improve the alignments such the model can capture that the simplification process involves in many cases splitting a sentence and then simplifying or train to learn when to split first and then attempt rewriting. 

The next two types of errors show failure in capturing discourse level meaning: a) errors due to failed pronoun resolution, shown in \ref{anaphoraresolution1} and \ref{anaphoraresolution2}, and b) errors due to the most important part of the sentence being left out, shown in \ref{wrongpart2} and \ref{wrongpart2}. In these cases, the sentences were not bad, but the information was assigned to the wrong referent, or important meaning was left out. In \ref{badsub1} and \ref{badsub2}, the substitution is clearly semantically related to the target, but changes the meaning. Finally, there were examples of acceptable simplifications, as in \ref{acceptable1} and \ref{acceptable2}, that were classified as errors because they were not in the gold data. We provide additional examples for each error category in the appendix.

To improve the performance of future models, we see several options. We can improve the original alignments within the Newsela corpus, particularly in the case where sentences are split. Prior to simplification, we can use additional context around the sentences to perform anaphora resolution; at this point, we can also learn when to perform sentence splitting; this has been done in the Hybrid model \cite{narayan2014hybrid}, but has not yet been incorporated into neural models. Finally, we can use syntactic information to ensure the main clause of a sentence is not removed.

\section{Conclusion}

In this paper, we present a novel Seq2Seq framework for sentence simplification. We contribute three major improvements over generic Seq2Seq models: a complexity-weighted loss function to encourage the model to choose simpler words; a similarity penalty during inference and clustering post-inference, to generate candidate simplifications with significant differences; and a reranking system to select the simplification that promotes both fluency and adequacy. Our model outperforms previous state-of-the-art systems using SARI, the standard metric for simplification. More importantly, while other previous models generate relatively long sentences, our model is able to generate shorter and simpler sentences, while remaining competitive regarding human-evaluated fluency and adequacy. Finally, we provide a qualitative analysis of where our different contributions improve performance, the effect of length on human-evaluated meaning preservation, and the current shortcomings of our model as insights for future research. 

Generating diverse outputs from Seq2Seq models could be used in a variety of NLP tasks, such as chatbots \cite{shao2017generating}, image captioning \cite{vijayakumar2018diverse}, and story generation \cite{fan2018hierarchical}. In addition, the proposed techniques can also be extremely helpful in leveled and personalized text simplification, where the goal is to generate different sentences based on who is requesting the simplification.

\section{Acknowledgments}

We would like to thank the anonymous reviewers for their helpful feedback on this work. We would also like to thank Devanshu Jain, Shyam Upadhyay, and Dan Roth for their feedback on the post-decoding aspect of this work, as well as Anne Cocos and Daphne Ippolito for their insightful comments during proofreading.

This  material  is  based  in  part  on  research  sponsored  by  DARPA  under  grant  number   HR0011-15-C-0115 (the LORELEI program). The U.S. Government is authorized to reproduce and distribute reprints for Governmental purposes. The views and conclusions contained in this publication are those of the authors and should not be interpreted as representing official policies or endorsements of DARPA and the U.S. Government.   

The  work  has  also  been  supported  by  the French National Research Agency under project ANR-16-CE33-0013. This research was partially supported by Jo\~{a}o Sedoc's Microsoft Research Dissertation Grant.
Finally, we gratefully acknowledge the support of NSF-SBIR grant 1456186.

\bibliography{naaclhlt2019}

\begin{thebibliography}{42}
\expandafter\ifx\csname natexlab\endcsname\relax\def\natexlab#1{#1}\fi

\bibitem[{Brants and Franz(2006)}]{thorsten2006web}
Thorsten Brants and Alex Franz. 2006.
\newblock Web 1t 5-gram version 1.
\newblock In \emph{LDC2006T13}, Philadelphia, Pennsylvania. Linguistic Data
  Consortium.

\bibitem[{Chandrasekar et~al.(1996)Chandrasekar, Doran, and
  Srinivas}]{chandrasekar1996motivations}
R.~Chandrasekar, Christine Doran, and B.~Srinivas. 1996.
\newblock Motivations and methods for text simplification.
\newblock In \emph{COLING 1996 Volume 2: The 16th International Conference on
  Computational Linguistics}.

\bibitem[{Chen et~al.(2015)Chen, Li, Li, Lin, Wang, Wang, Xiao, Xu, Zhang, and
  Zhang}]{chen2015mxnet}
Tianqi Chen, Mu~Li, Yutian Li, Min Lin, Naiyan Wang, Minjie Wang, Tianjun Xiao,
  Bing Xu, Chiyuan Zhang, and Zheng Zhang. 2015.
\newblock {MXNet: A Flexible and Efficient Machine Learning Library for
  Heterogeneous Distributed Systems}.
\newblock \emph{CoRR}, abs/1512.01274.

\bibitem[{Coster and Kauchak(2011)}]{coster2011learning}
Will Coster and David Kauchak. 2011.
\newblock Learning to simplify sentences using wikipedia.
\newblock In \emph{Proceedings of the Workshop on Monolingual Text-To-Text
  Generation}, pages 1--9, Portland, Oregon. Association for Computational
  Linguistics.

\bibitem[{Fan et~al.(2018)Fan, Lewis, and Dauphin}]{fan2018hierarchical}
Angela Fan, Mike Lewis, and Yann Dauphin. 2018.
\newblock Hierarchical neural story generation.
\newblock In \emph{Proceedings of the 56th Annual Meeting of the Association
  for Computational Linguistics (Volume 1: Long Papers)}, pages 889--898,
  Melbourne, Australia. Association for Computational Linguistics.

\bibitem[{Heafield(2011)}]{heafield2011kenlm}
Kenneth Heafield. 2011.
\newblock {KenLM:} faster and smaller language model queries.
\newblock In \emph{Proceedings of the {EMNLP} 2011 Sixth Workshop on
  Statistical Machine Translation}, pages 187--197, Edinburgh, Scotland, UK.

\bibitem[{Hieber et~al.(2017)Hieber, Domhan, Denkowski, Vilar, Sokolov,
  Clifton, and Post}]{hieber2017sockeye}
Felix Hieber, Tobias Domhan, Michael Denkowski, David Vilar, Artem Sokolov, Ann
  Clifton, and Matt Post. 2017.
\newblock Sockeye: A toolkit for neural machine translation.
\newblock \emph{CoRR}, abs/1712.05690.

\bibitem[{Kim(2014)}]{kim2014convolutional}
Yoon Kim. 2014.
\newblock Convolutional neural networks for sentence classification.
\newblock In \emph{Proceedings of the 2014 Conference on Empirical Methods in
  Natural Language Processing (EMNLP)}, pages 1746--1751, Doha, Qatar.
  Association for Computational Linguistics.

\bibitem[{Kincaid et~al.(1975)Kincaid, Fishburne, Rogers, and
  Chissom}]{kincaid1975derivation}
J.~Peter Kincaid, Robert~P. Fishburne, Richard E.~L. Rogers, and Brad~S.
  Chissom. 1975.
\newblock Derivation of new readability formulas (automated readability index,
  fog count and flesch reading ease formula) for navy enlisted personnel ;
  research branch report 8-75.

\bibitem[{Kingma and Ba(2015)}]{kingma2015adam}
Diederik~P. Kingma and Jimmy Ba. 2015.
\newblock Adam: A method for stochastic optimization.
\newblock \emph{International Conference on Learning Representations}.

\bibitem[{Kriz et~al.(2018)Kriz, Miltsakaki, Apidianaki, and
  Callison-Burch}]{kriz2018simplification}
Reno Kriz, Eleni Miltsakaki, Marianna Apidianaki, and Chris Callison-Burch.
  2018.
\newblock Simplification using paraphrases and context-based lexical
  substitution.
\newblock In \emph{Proceedings of the 2018 Conference of the North American
  Chapter of the Association for Computational Linguistics: Human Language
  Technologies, Volume 1 (Long Papers)}, pages 207--217, New Orleans,
  Louisiana. Association for Computational Linguistics.

\bibitem[{Le and Mikolov(2014)}]{le2014distributed}
Quoc Le and Tomas Mikolov. 2014.
\newblock Distributed representations of sentences and documents.
\newblock In \emph{Proceedings of the 31st International Conference on
  International Conference on Machine Learning - Volume 32}, ICML'14, pages
  1188--1196. JMLR.org.

\bibitem[{Li et~al.(2016)Li, Monroe, and Jurafsky}]{li2016simple}
Jiwei Li, Will Monroe, and Dan Jurafsky. 2016.
\newblock {A Simple, Fast Diverse Decoding Algorithm for Neural Generation}.
\newblock \emph{CoRR}.

\bibitem[{Luong et~al.(2015)Luong, Pham, and Manning}]{luong2015effective}
Thang Luong, Hieu Pham, and Christopher~D. Manning. 2015.
\newblock Effective approaches to attention-based neural machine translation.
\newblock In \emph{Proceedings of the 2015 Conference on Empirical Methods in
  Natural Language Processing}, pages 1412--1421, Lisbon, Portugal. Association
  for Computational Linguistics.

\bibitem[{Manning et~al.(2014)Manning, Surdeanu, Bauer, Finkel, Bethard, and
  McClosky}]{manning2014stanford}
Christopher Manning, Mihai Surdeanu, John Bauer, Jenny Finkel, Steven Bethard,
  and David McClosky. 2014.
\newblock The stanford corenlp natural language processing toolkit.
\newblock In \emph{Proceedings of 52nd Annual Meeting of the Association for
  Computational Linguistics: System Demonstrations}, pages 55--60, Baltimore,
  Maryland. Association for Computational Linguistics.

\bibitem[{Mikolov et~al.(2013)Mikolov, Sutskever, Chen, Corrado, and
  Dean}]{mikolov2013distributed}
Tomas Mikolov, Ilya Sutskever, Kai Chen, Greg~S Corrado, and Jeff Dean. 2013.
\newblock {Distributed Representations of Words and Phrases and their
  Compositionality}.
\newblock In \emph{Neural Information Processing Systems}, pages 3111--3119,
  Lake Tahoe, Nevada.

\bibitem[{Nallapati et~al.(2016)Nallapati, Zhou, dos Santos, Gulcehre, and
  Xiang}]{nallapati2016abstractive}
Ramesh Nallapati, Bowen Zhou, Cicero dos Santos, Caglar Gulcehre, and Bing
  Xiang. 2016.
\newblock Abstractive text summarization using sequence-to-sequence rnns and
  beyond.
\newblock In \emph{Proceedings of The 20th SIGNLL Conference on Computational
  Natural Language Learning}, pages 280--290, Berlin, Germany. Association for
  Computational Linguistics.

\bibitem[{Napoles et~al.(2011)Napoles, Van~Durme, and
  Callison-Burch}]{napoles2011evaluating}
Courtney Napoles, Benjamin Van~Durme, and Chris Callison-Burch. 2011.
\newblock Evaluating sentence compression: Pitfalls and suggested remedies.
\newblock In \emph{Proceedings of the Workshop on Monolingual Text-To-Text
  Generation}, pages 91--97, Portland, Oregon. Association for Computational
  Linguistics.

\bibitem[{Narayan and Gardent(2014)}]{narayan2014hybrid}
Shashi Narayan and Claire Gardent. 2014.
\newblock Hybrid simplification using deep semantics and machine translation.
\newblock In \emph{Proceedings of the 52nd Annual Meeting of the Association
  for Computational Linguistics (Volume 1: Long Papers)}, pages 435--445,
  Baltimore, Maryland. Association for Computational Linguistics.

\bibitem[{Nisioi et~al.(2017)Nisioi, {\v{S}}tajner, Ponzetto, and
  Dinu}]{nisioi2017exploring}
Sergiu Nisioi, Sanja {\v{S}}tajner, Simone~Paolo Ponzetto, and Liviu~P. Dinu.
  2017.
\newblock Exploring neural text simplification models.
\newblock In \emph{Proceedings of the 55th Annual Meeting of the Association
  for Computational Linguistics (Volume 2: Short Papers)}, pages 85--91,
  Vancouver, Canada. Association for Computational Linguistics.

\bibitem[{Papineni et~al.(2002)Papineni, Roukos, Ward, and
  Zhu}]{papineni2002bleu}
Kishore Papineni, Salim Roukos, Todd Ward, and Wei-Jing Zhu. 2002.
\newblock Bleu: a method for automatic evaluation of machine translation.
\newblock In \emph{Proceedings of 40th Annual Meeting of the Association for
  Computational Linguistics}, pages 311--318, Philadelphia, Pennsylvania, USA.
  Association for Computational Linguistics.

\bibitem[{Parker et~al.(2011)Parker, Graff, Kong, Chen, and
  Maeda}]{parker2011english}
Robert Parker, David Graff, Junbo Kong, Ke~Chen, and Kazuaki Maeda. 2011.
\newblock {English Gigaword Fifth Edition LDC2011T07. DVD.}
\newblock \emph{Philadelphia: Linguistic Data Consortium}.

\bibitem[{Pennington et~al.(2014)Pennington, Socher, and
  Manning}]{pennington2014glove}
Jeffrey Pennington, Richard Socher, and Christopher Manning. 2014.
\newblock Glove: Global vectors for word representation.
\newblock In \emph{Proceedings of the 2014 Conference on Empirical Methods in
  Natural Language Processing (EMNLP)}, pages 1532--1543, Doha, Qatar.
  Association for Computational Linguistics.

\bibitem[{Saggion(2017)}]{saggion2017automatic}
Horacio Saggion. 2017.
\newblock \emph{Automatic Text Simplification}.
\newblock Synthesis Lectures on Human Language Technologies. Morgan {\&}
  Claypool Publishers.

\bibitem[{Shao et~al.(2017)Shao, Gouws, Britz, Goldie, Strope, and
  Kurzweil}]{shao2017generating}
Louis Shao, Stephan Gouws, Denny Britz, Anna Goldie, Brian Strope, and Ray
  Kurzweil. 2017.
\newblock Generating high-quality and informative conversation responses with
  sequence-to-sequence models.
\newblock In \emph{Proceedings of the 2017 Conference on Empirical Methods in
  Natural Language Processing}, pages 2210--2219.

\bibitem[{Snover et~al.(2006)Snover, Dorr, Schwartz, Micciulla, and
  Makhoul}]{snover2006study}
Matthew Snover, Bonnie Dorr, Richard Schwartz, Linnea Micciulla, and John
  Makhoul. 2006.
\newblock {A Study of Translation Edit Rate with Targeted Human Annotation}.
\newblock In \emph{Proceedings of Association for Machine Translation in the
  Americas}, pages 223--231, Cambridge, MA.

\bibitem[{{\v{S}}tajner and Popovic(2016)}]{stajner2016text}
Sanja {\v{S}}tajner and Maja Popovic. 2016.
\newblock Can text simplification help machine translation?
\newblock In \emph{Proceedings of the 19th Annual Conference of the European
  Association for Machine Translation}, pages 230--242.

\bibitem[{Sulem et~al.(2018)Sulem, Abend, and Rappoport}]{sulem2018bleu}
Elior Sulem, Omri Abend, and Ari Rappoport. 2018.
\newblock Bleu is not suitable for the evaluation of text simplification.
\newblock In \emph{Proceedings of the 2018 Conference on Empirical Methods in
  Natural Language Processing}, pages 738--744, Brussels, Belgium. Association
  for Computational Linguistics.

\bibitem[{Sutskever et~al.(2014)Sutskever, Vinyals, and
  Le}]{sutskever2014sequence}
Ilya Sutskever, Oriol Vinyals, and Quoc~V. Le. 2014.
\newblock {Sequence to Sequence Learning with Neural Networks}.
\newblock In \emph{Neural Information Processing Systems}, Montreal, Canada.

\bibitem[{Vaswani et~al.(2017)Vaswani, Shazeer, Parmar, Uszkoreit, Jones,
  Gomez, Kaiser, and Polosukhin}]{vaswani2017attention}
Ashish Vaswani, Noam Shazeer, Niki Parmar, Jakob Uszkoreit, Llion Jones,
  Aidan~N. Gomez, Lukasz Kaiser, and Illia Polosukhin. 2017.
\newblock {Attention Is All You Need}.
\newblock In \emph{Neural Information Processing Systems}, Long Beach, CA.

\bibitem[{Vickrey and Koller(2008)}]{vickrey2008sentence}
David Vickrey and Daphne Koller. 2008.
\newblock Sentence simplification for semantic role labeling.
\newblock In \emph{Proceedings of ACL-08: HLT}, Columbus, Ohio. Association for
  Computational Linguistics.

\bibitem[{Vijayakumar et~al.(2018)Vijayakumar, Cogswell, Selvaraju, Sun, Lee,
  Crandall, and Batra}]{vijayakumar2018diverse}
Ashwin~K Vijayakumar, Michael Cogswell, Ramprasath~R Selvaraju, Qing Sun,
  Stefan Lee, David Crandall, and Dhruv Batra. 2018.
\newblock Diverse beam search: Decoding diverse solutions from neural sequence
  models.
\newblock In \emph{AAAI Conference on Artificial Intelligence (AAAI)}.

\bibitem[{Vinyals and Le(2015)}]{vinyals2015neural}
Oriol Vinyals and Quoc~V. Le. 2015.
\newblock A neural conversational model.
\newblock In \emph{Proceedings of the International Conference on Machine
  Learning, Deep Learning Workshop}.

\bibitem[{Vu et~al.(2018)Vu, Hu, Munkhdalai, and Yu}]{vu2018sentence}
Tu~Vu, Baotian Hu, Tsendsuren Munkhdalai, and Hong Yu. 2018.
\newblock Sentence simplification with memory-augmented neural networks.
\newblock In \emph{Proceedings of the 2018 Conference of the North American
  Chapter of the Association for Computational Linguistics: Human Language
  Technologies, Volume 2 (Short Papers)}, pages 79--85.

\bibitem[{Woodsend and Lapata(2011)}]{woodsend2011learning}
Kristian Woodsend and Mirella Lapata. 2011.
\newblock Learning to simplify sentences with quasi-synchronous grammar and
  integer programming.
\newblock In \emph{Proceedings of the 2011 Conference on Empirical Methods in
  Natural Language Processing}, pages 409--420, Edinburgh, Scotland, UK.
  Association for Computational Linguistics.

\bibitem[{Woodsend and Lapata(2014)}]{woodsend2014text}
Kristian Woodsend and Mirella Lapata. 2014.
\newblock Text rewriting improves semantic role labeling.
\newblock \emph{Journal of Artificial Intelligence Research}, 51:133--164.

\bibitem[{Wubben et~al.(2012)Wubben, van~den Bosch, and
  Krahmer}]{wubben2012sentence}
Sander Wubben, Antal van~den Bosch, and Emiel Krahmer. 2012.
\newblock Sentence simplification by monolingual machine translation.
\newblock In \emph{Proceedings of the 50th Annual Meeting of the Association
  for Computational Linguistics (Volume 1: Long Papers)}, pages 1015--1024,
  Jeju Island, Korea. Association for Computational Linguistics.

\bibitem[{Xu et~al.(2015)Xu, Callison-Burch, and Napoles}]{xu2015problems}
Wei Xu, Chris Callison-Burch, and Courtney Napoles. 2015.
\newblock Problems in current text simplification research: New data can help.
\newblock \emph{Transactions of the Association for Computational Linguistics},
  3(1):283--297.

\bibitem[{Xu et~al.(2016)Xu, Napoles, Pavlick, Chen, and
  Callison-Burch}]{xu2016optimizing}
Wei Xu, Courtney Napoles, Ellie Pavlick, Quanze Chen, and Chris Callison-Burch.
  2016.
\newblock Optimizing statistical machine translation for text simplification.
\newblock \emph{Transactions of the Association for Computational Linguistics},
  4(1):401--415.

\bibitem[{Zhang and Lapata(2017)}]{zhang2017sentence}
Xingxing Zhang and Mirella Lapata. 2017.
\newblock Sentence simplification with deep reinforcement learning.
\newblock In \emph{Proceedings of the 2017 Conference on Empirical Methods in
  Natural Language Processing}, pages 584--594. Association for Computational
  Linguistics.

\bibitem[{Zhao et~al.(2018)Zhao, Meng, He, Andi, and
  Bambang}]{zhao2018integrating}
Sanqiang Zhao, Rui Meng, Daqing He, Saptono Andi, and Parmanto Bambang. 2018.
\newblock Integrating transformer and paraphrase rules for sentence
  simplification.
\newblock In \emph{Proceedings of the 2018 EMNLP Conference}, pages 3164--3173,
  Brussels, Belgium.

\bibitem[{Zhu et~al.(2010)Zhu, Bernhard, and Gurevych}]{zhu2010monolingual}
Zhemin Zhu, Delphine Bernhard, and Iryna Gurevych. 2010.
\newblock A monolingual tree-based translation model for sentence
  simplification.
\newblock In \emph{Proceedings of the 23rd International Conference on
  Computational Linguistics (Coling 2010)}, pages 1353--1361, Beijing, China.

\end{thebibliography}
\bibliographystyle{acl_natbib}

\end{document}